\documentclass{article}

\usepackage[dblblindworkshop, final]{neurips_2025}

\usepackage[utf8]{inputenc} 
\usepackage[T1]{fontenc}    
\usepackage{hyperref}       
\usepackage{url}            
\usepackage{booktabs}       
\usepackage{amsfonts}       
\usepackage{nicefrac}       
\usepackage{microtype}      
\usepackage{xcolor}         
\usepackage{amsmath}
\usepackage{multirow}       
\usepackage{wrapfig}        
\usepackage{graphicx}
\usepackage{subcaption}
\usepackage{enumitem}       
\usepackage{dsfont}
\usepackage{placeins}

\workshoptitle{Efficient Reasoning}

\title{Influence Functions for  Efficient Data Selection in Reasoning}

\author{
Prateek Humane$^{1,2, \dagger}$ \quad
Paolo Cudrano$^{3}$ \quad
Daniel Z. Kaplan$^{4}$ \\
\textbf{Matteo Matteucci}$^{3}$ \quad
\textbf{Supriyo Chakraborty}$^{5}$ \quad
\textbf{Irina Rish}$^{1,2}$ \quad
\\ \\
$^{1}$Mila, Québec AI Institute,
$^{2}$Université de Montréal, 
$^{3}$Politecnico di Milano, \\
$^{4}$Sage Group,
$^{5}$Capital One
\\ 
$^\dagger$\texttt{prateek.humane@mila.quebec}
}

\begin{document}

\maketitle

 

\begin{abstract}
Fine-tuning large language models (LLMs) on chain-of-thought (CoT) data shows that a small amount of high-quality data can outperform massive datasets. Yet, what constitutes “quality” remains ill-defined. Existing reasoning methods rely on indirect heuristics such as problem difficulty or trace length, while instruction-tuning has explored a broader range of automated selection strategies—but rarely in the context of reasoning. We propose to define reasoning data quality using influence functions, which measure the causal effect of individual CoT examples on downstream accuracy, and introduce influence-based pruning, which consistently outperforms perplexity and embedding-based baselines on math reasoning within a model family.
\footnote{Code available at: https://github.com/mila-capitalone-collab/inf-functions-reasoning/tree/main}
\end{abstract}

\section{Introduction}
Large language models (LLMs) show strong reasoning gains when fine-tuned on chain-of-thought (CoT) data \citep{deepseekai2025deepseekr1incentivizingreasoningcapability}. Recent studies suggest that data quality often outweighs quantity \citep{rodchenko2025aiproblemdataproblem,yuan2025naturalreasoningreasoningwild28m, ye2025limoreasoning}, yet what constitutes “quality” remains unclear.

Prior work on reasoning has used proxies for data quality, such as question difficulty \citep{chen2025advancingmathematicalreasoninglanguage} or CoT length \citep{yuan2025naturalreasoningreasoningwild28m}, which offer only indirect signals of downstream performance. In contrast, instruction-tuning studies have systematically evaluated automated data selection methods—perplexity filtering, embedding similarity, gradient-based scoring \citep{yin2025computeconstraineddataselection, rds}—but these have not been fully applied to reasoning data.

What ultimately matters is not whether an example appears challenging or diverse, but whether fine-tuning on it {\em causally improves reasoning performance}. Influence functions (IFs) provide a principled way to capture such causal effects by estimating the contribution of individual training examples to downstream accuracy. While IFs have been explored in instruction tuning and knowledge tracing, they remain unused for reasoning-specific pruning.

In this work, we extend IFs to define reasoning data quality through causal impact on downstream accuracy. We introduce IF-based pruning, which surpasses common baselines on math reasoning within a model family. However, transfer across families remains inconclusive, raising the open question of whether data quality is intrinsic or model-specific.

\section{Related Works}

Let $\mathcal{D}$ denote a training dataset and $\mathcal{V}$ a validation set sampled from the evaluation distribution. Data selection methods assign scores to examples in $\mathcal{D}$ in order to prune/reweight the training pool. These methods either (1) score examples directly, $s: \mathcal{D} \to \mathbb{R}$, or (2) compute pairwise scores with respect to $\mathcal{V}$, $s: \mathcal{V} \times \mathcal{D} \to \mathbb{R}$, then aggregate across $v \in \mathcal{V}$ to obtain $s(d)$ for each $d \in \mathcal{D}$.
We group existing approaches accordingly.

\paragraph{Reasoning heuristics} 

Heuristic methods define scores $s(d)$ directly from inherent properties of each training example $d \in \mathcal{D}$, independent of $\mathcal{V}$ or a specific model. Recent works take into account heuristics such as estimated question difficulty, CoT length, and problem diversity \citep{ye2025limoreasoning, s1}. 
LIMO \citep{ye2025limoreasoning} also incorporates human evaluation, selecting solutions that exhibit desirable behaviors such as elaboration and self-verification. Select2Reason \citep{select2reason} provides a systematic study, showing that higher $s(d)$ assigned to examples with longer traces and harder problems yields stronger downstream performance, whereas diversity has limited benefit. 


\paragraph{Perplexity based} These methods score each example $s(d)$ by its negative log-likelihood under the base model. \textit{Top-PPL} selects high-perplexity (hard) cases, while \textit{Mid-PPL} favors mid-range examples to avoid trivial or outlier data. These scores are cheap to compute and have shown to improve sample efficiency in the instruction tuning setting \citep{yin2025computeconstraineddataselection, ankner2024perplexedperplexityperplexitybaseddata}.
\paragraph{Embedding based} Embedding-based methods use data representations to assign scores. 
S1 \citep{s1} enforces coverage by classifying problems into topics with an LLM and sampling across them. RDS+ \citep{rds} computes pairwise scores $s(v,d) = \langle E(v), E(d) \rangle$ between validation and training examples, 
selecting examples most relavent to each $v \in \mathcal{V}$ in a round-robin fashion.
\paragraph{Gradient based}
Gradient and influence based methods compute $s(v,d)$ by estimating the causal contribution of training examples on validation loss.
TracIn \citep{tracin} sums gradient inner products across training checkpoints and LESS \citep{LESS} scales this approach to large models using gradient datastores. InfDist \citep{infdist} learns optimal per-example weights so that a single gradient step minimizes validation loss.
\paragraph{Our influence function approach}
Unlike gradient-similarity methods that track training dynamics, influence functions measure the effect of a training example on converged model parameters and downstream objectives
$f(\theta^\star)$. 
Formally, the influence of a training example $d \in \mathcal{D}$ is
$
\mathcal{I}_f(d) 
= - \nabla_\theta f(\theta^\star)^\top H^{-1} \nabla_\theta L(d, \theta^\star)
$
where $H$ is the Hessian of the training objective at $\theta^\star$. Approximations such as EK-FAC have made this tractable for LLMs \citep{grosse2023studyinglargelanguagemodel}. \citet{ ruis2025proceduralknowledgepretrainingdrives} shows that examples with high estimated influence on cross-entropy loss also causally affect downstream accuracy, and their removal causes larger performance drops than random pruning or TracIn.

We extend this idea to reasoning, scoring examples by their influence on whether fine-tuning improves or harms correctness. This approach yields scores $s(d)$ that capture how each training example $d \in \mathcal{D}$ steers the model toward or away from correct reasoning behavior.




\section{Methods}
\subsection{Influence Functions Scoring}
Let $\theta$ be a pretrained model and $\mathcal{D}$ the LIMO training dataset \citep{ye2025limoreasoning}. We fine-tune $\theta$ on $\mathcal{D}$ for 10 epochs following the settings of \citet{ye2025limoreasoning}, obtaining $\theta^{\text{LIMO}}$. We then evaluate on the MATH500 benchmark $\mathcal{V}$ for both model parameters $\theta$ and $\theta^\text{LIMO}$ \citep{hendrycks2021measuring}. 

To analyze how fine-tuning changes behavior, we compare correctness before and after fine-tuning for each query $(q,a) \in \mathcal{V}$. This yields two disjoint subsets $C,I \subseteq \mathcal{V}$:
\begin{align*}
C = \{(q,a) \in \mathcal{V}: \text{Acc}(f(q;\theta),a)=0 \land \text{Acc}(f(q;\theta^\text{LIMO}),a)=1\}\\
I = \{(q,a) \in \mathcal{V}: \text{Acc}(f(q;\theta),a)=1 \land \text{Acc}(f(q;\theta^\text{LIMO}),a)=0\}
\end{align*}
Here $\text{Acc}(f(q;\theta),a)$ is a binary indicator of whether the model $f(\cdot;\theta)$ greedy sampling completion for query $q$ matches the ground-truth answer $a$. Thus, $C$ collects queries where fine-tuning improves correctness, while $I$ collects queries where fine-tuning degrades correctness.

Our goal is to estimate the causal influence of training examples $d \in \mathcal{D}$ on these outcome sets. Because influence functions require a differentiable proxy, we follow \citet{grosse2023studyinglargelanguagemodel} and compute influence with respect to cross-entropy loss on completions. For each validation query $v \in C \cup I$ with completion $y_c$, the influence of a training point $d$ is:
\[
\mathcal{I}(d;(q,a)) 
= - \nabla_{\theta} {\mathcal{L}_\text{CE}(f(q;\theta^\text{LIMO}),\theta)}^\top H^{-1} \nabla_\theta \mathcal{L}_\text{CE}(d, \theta^\text{LIMO})
\]
where ${\mathcal{L}_\text{CE}(f(q;\theta^\text{LIMO}),\theta)}$ is the cross-entropy loss on the generated completion and $\mathcal{L}_\text{CE}(d, \theta^\text{LIMO})$ is the training loss of $d$ at convergence.

By aggregating $\mathcal{I}(d;v)$ across queries $v \in C$ and $v \in I$, we obtain scores $s(d)$ that quantify whether training on $d$ tends to push the model toward correct completions (beneficial influence) or toward incorrect completions (harmful influence). These scores provide a principled, influence-based ranking of reasoning examples in $\mathcal{D}$.

\begin{wrapfigure}{r}{0.5\textwidth}
  \begin{center}
 \vspace{-20pt}
    \includegraphics[width=0.49\textwidth]{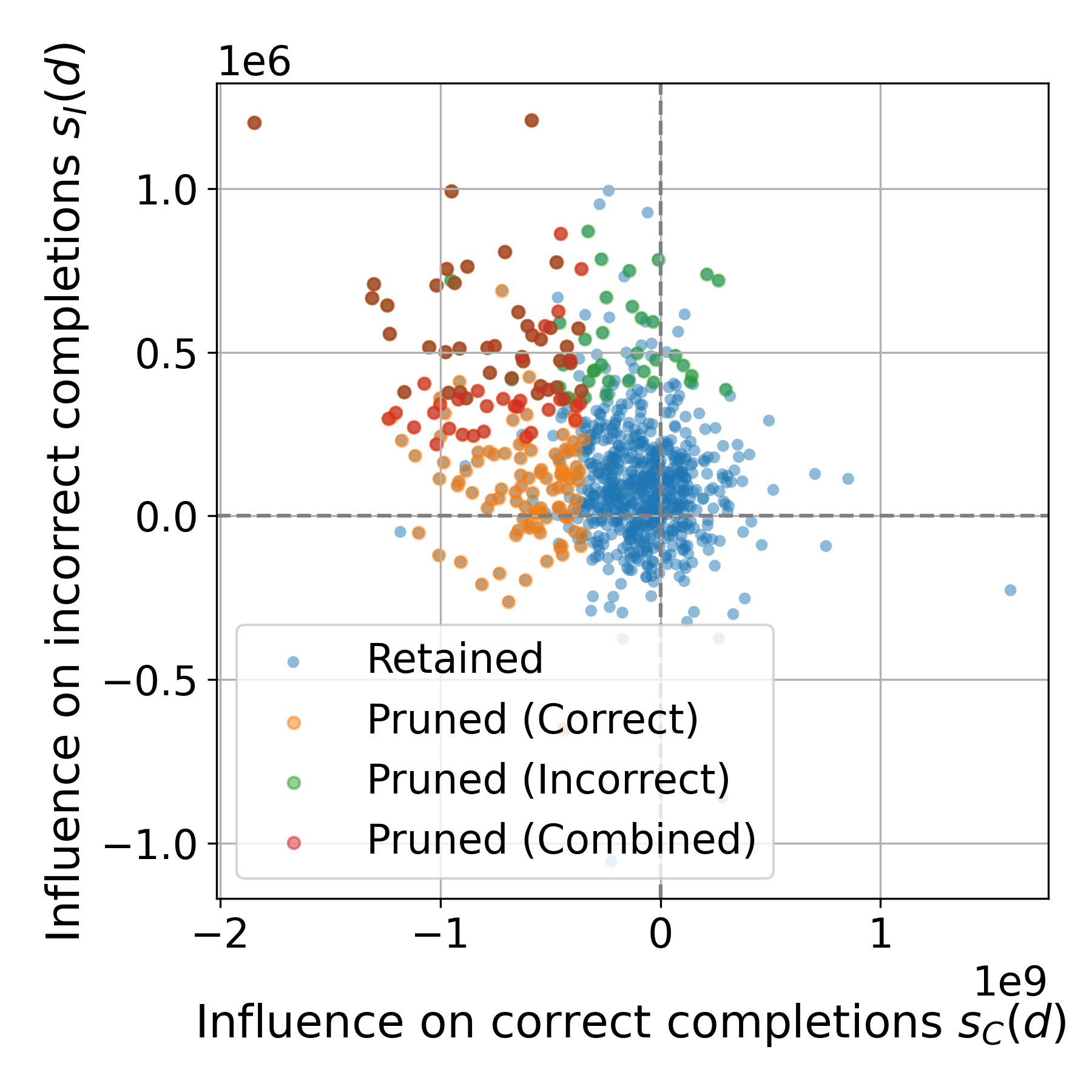}
  \end{center}
 \vspace{-10pt}
\caption{Influence scores $s_C(d), s_I(d)$ for training examples $d \in \mathcal{D}$. Points farther left contribute less to correct answers, while points higher up push the model toward wrong answers. Interestingly, the majority of outliers appear in the top-left quadrant, where lie the datapoints we predict to be most harmful. \textit{Note: as the final pruning is affected also by rank scores $r(d)$, few unpruned points may also be found in such quadrant.}}
\label{fig:scatter_main}
 \vspace{-4em}
\end{wrapfigure}
 
\subsection{Data Pruning}






We first aggregate influence scores across validation queries $v$:

\resizebox{\linewidth}{!}{%
    \begin{minipage}{\linewidth}\vspace{-1em}
\begin{align*}
s_C(d) &= \frac{1}{|C|}\sum_{v\in{C}}  \mathcal{I}(d;v), &s_I(d) = \frac{1}{|I|}\sum_{v\in{I}}  \mathcal{I}(d;v).\\
\end{align*}
\end{minipage}%
}\vspace{-1.3em}

To prevent bias from queries with disproportionately large scores, we complement raw influence scores $s(d)$ with rank-based measures $r(d)$. For each query $v$, we sort training examples by their influence values  in descending order and assign ranks. The final rank for a datapoint is its average position across queries:

\resizebox{\linewidth}{!}{%
    \begin{minipage}{\linewidth}\vspace{-1em}
\begin{align*}
 r_C(d) &= \frac{1}{|C|}\sum_{v\in{C}}  \text{rank}(\mathcal{I}(d;v)), 
 &r_I(d) = \frac{1}{|I|}\sum_{v\in{I}}  \text{rank}(\mathcal{I}(d;v)).
\end{align*}
\end{minipage}%
}

Pruning is then performed by \textbf{intersecting thresholds on both scores and ranks}, ensuring that selected datapoints are consistently high or low influence across queries, rather than dominated by outliers.

We evaluate three pruning strategies (See Figure \ref{fig:scatter_main}). \textbf{(\romannumeral 1) \emph{Correct}}: prune the bottom 20\% of examples by intersecting low $s_C(d)$ and $r_C(d)$, removing examples that contribute least to correct completions; \textbf{(\romannumeral 2) \emph{Incorrect}}: prune the top 10\% of examples by intersecting high $s_I(d)$ and $r_I(d)$, removing examples that most strongly push the model toward incorrect completions; \textbf{(\romannumeral 3) \emph{Combined}}: prune 10\% of examples that satisfy both of the above criteria.

Interestingly, histograms of the score distributions in Figure~\ref{fig:histograms} show long tails: in the correct case, many examples have negligible or negative benefit, while in the incorrect case, outliers with high influence dominate harmful behavior.

\section{Experiments and Results}
We perform supervised fine-tuning (SFT) on pruned versions of the LIMO dataset, using two models: \mbox{LLaMA-3-8B-Instruct}~\citep{grattafiori2024llama}, also used during pruning, and \mbox{Qwen2.5-Math-7B-Instruct}~\citep{yang2024qwen2}, external to the pruning process. We ran full finetuning following the training settings from LIMO for 10 epochs, saving checkpoints at every other epoch. Given that we already were optimizing for MATH500, we treated it as a validation set and chose the checkpoint which performed best on MATH500. We chose this strategy as for different datasets, optimal checkpoints were not consistent.

Evaluation focuses on the math reasoning benchmarks: AIME24, AMC23, OlympiadBench~\citep{he2024olymbpiadbench}, GSM8k~\citep{cobbe2021training}. Additionally we include GPQA~\citep{rein2024gpqa} to assess transfer to non-math reasoning. 
We compare three IF-based pruning strategies (\emph{Correct}, \emph{Incorrect}, \emph{Combined}) against \emph{Random}, \emph{Mid-PPL}~\cite{}, and \emph{RDS+}~\citep{rds}, and report also the unpruned dataset and the base model for reference. 
All results use the unbiased pass@1 metric~\citep{chen2021evaluating} with $N{=}8$ samples in a zero-shot chain-of-thought setting, following the LIMO protocol. 
Full training and evaluation details are provided in the Appendix.


\begin{description}[leftmargin=0cm]
  \item[Within-model IF pruning is effective.] 
  On Llama3-8B-Instruct (Table~\ref{tab:main_res_llama}), our IF-selected subsets match or outperform \emph{Random}/\emph{Mid-PPL} baselines and \emph{RDS+}.
  Results on AIME24 remain neglibigle across all methods, reflecting the benchmark’s difficulty and indicating that SFT on small pruned subsets cannot recover performance.
  On GPQA, math SFT tends to degrade general scientific reasoning; our selections mitigate but do not eliminate this effect.
  Even when pruning more aggressively 50\% of the data, IF-based selection improves upon or at least remains competitive with baselines (see Appendix, Table~\ref{tab:res_50percent_llama}).
  \item[Cross-family transfer is non-conclusive.]
  When fine-tuning Qwen2.5-Math-7B-Instruct with data subsets selected using Llama3-8B-Instruct (Appendix, Table~\ref{tab:main_res_qwen}),
  winners vary by task and no pruning strategy consistently outperforms the baselines. This suggests that improvements from subsets selected with one model do not straightforwardly carry over to a different model family. 
\end{description}

\begin{table}[t]
\centering
\caption{\textbf{Results on \mbox{LLaMA-3-8B-Instruct}.} 
Pass@1 with $N{=}8$ samples, zero-shot CoT. 
We fine-tune on math-pruned subsets selected by influence functions (\emph{Correct}, \emph{Incorrect}, \emph{Combined}) and compare against baselines (\emph{Random}, \emph{Mid-PPL}, \emph{RDS+}), as well as the unpruned dataset and base model. 
Evaluation covers math reasoning benchmarks (AIME24, AMC23, OlympiadBench, GSM8k) and GPQA for non-math transfer. Best in \textbf{bold}; best-after-SFT \underline{underlined}. Amount of samples not pruned in [brackets].}
\label{tab:main_res_llama}
\vspace{.5em}
\footnotesize
\begin{tabular}{@{}lccccc@{}}
\multicolumn{6}{l}{\textsc{Llama3-8B-Instruct}}\\
\toprule
\textsc{Pruning} & \textsc{GSM8k} & \textsc{OlympiadBench} & \textsc{AMC23} & \textsc{AIME24} & \textsc{GPQA} \\
\midrule
Combined [90\%] (ours)   & \textbf{0.8113} & \textbf{0.0880} & \textbf{0.1156} & 0.0000 & \underline{0.1654} \\
Incorrect [90\%] (ours)  & 0.7878 & 0.0798 & 0.1094 & 0.0042 & 0.1439 \\
Correct [80\%] (ours)    & 0.7922 & 0.0691 & 0.0719 & \textbf{0.0083} & 0.1496 \\
\cmidrule{1-6}
RDS+ [90\%]       & 0.8015 & \textbf{0.0880} & 0.0813 & 0.0042 & 0.1553 \\
Mid-PPL [90\%]    & 0.8052 & 0.0796 & 0.0781 & 0.0000 & 0.1244 \\
Random [90\%]     & 0.7839 & 0.0865 & 0.1094 & 0.0042 & 0.1509 \\
None       & 0.7722 & 0.0685 & 0.0813 & 0.0000 & 0.1199 \\
\cmidrule{1-6}
W/o SFT    & 0.5958 & 0.0498 & 0.0781 & 0.0000 & \textbf{0.2973} \\
\bottomrule
\end{tabular}
\end{table}

\section{Limitations and Future Work}

While our study demonstrates the promise of influence-function–based pruning for reasoning, several limitations and opportunities for future work remain.

\textbf{Experimental variance.} Our results are based on a single training run per setting. Future work should include multiple seeds and confidence intervals to account for evaluation variance.

\textbf{Dataset scope.} We focus on LIMO, which is already pruned and curated for questions \mbox{Qwen2.5-Math-7B-Instruct} cannot solve. This may underestimate the potential of influence functions on larger, less filtered datasets such as Open-R1.

\textbf{Comprehensive evaluation.} We evaluate only a limited set of baselines and a single model size. We are planning a comprehensive study comparing influence-based pruning against  reasoning heuristics and other methods such as InfDist, while spanning different datasets and model scales, to better understand when each approach is effective.

\textbf{Computational cost.} Influence-function estimation is expensive, and we do not account for the additional compute required for data selection. Future work should explore more efficient approximations, such as gradient datastores or training LLMs to predict influence scores directly.

\textbf{Evolving data quality.} Finally, we treat data quality as static, but in practice it may evolve as models improve. An open question is how influence scores shift with scale and training stage—for example, whether higher-quality examples track increasingly difficult problems as models become stronger.

\bibliographystyle{apalike}
\bibliography{refs}

\FloatBarrier
\appendix

\section{Training and Evaluation Details}

\paragraph{Training.} Our SFT adopts the same training setup as LIMO \citep{ye2025limoreasoning}.
All experiments are run on 8$\times$H100 GPUs, with full-parameter fine-tuning for 10 epochs. We use a cosine learning rate schedule with peak learning rate $5 \times 10^{-6}$.

\paragraph{Evaluation.} We follow the evaluation setup of LIMO~\citep{ye2025limoreasoning}, with the addition of the GSM8k test set manually added.
We report the unbiased pass@1 metric \citep{chen2021evaluating}, estimated from $N{=}8$ sampled generations.
Sampling uses temperature $0.6$, top-$p=0.95$, top-$k=1$, and a maximum generation length of $16{,}384$ tokens.

\section{Notes on influence functions computation}

To compute influence function scores, we follow the derivation and approximations described in \citet{grosse2023studyinglargelanguagemodel}.
Following \citet{ruis2025proceduralknowledgepretrainingdrives},we perform the computation considering only MLP layers.

\section{Further details on our pruning strategies}

\begin{figure}[ht]
\centering
\begin{minipage}[b]{0.49\textwidth}
\begin{subfigure}{\textwidth}
    \caption{\emph{Correct} influence score $s_C(d)$.}
    \includegraphics[width=\textwidth]{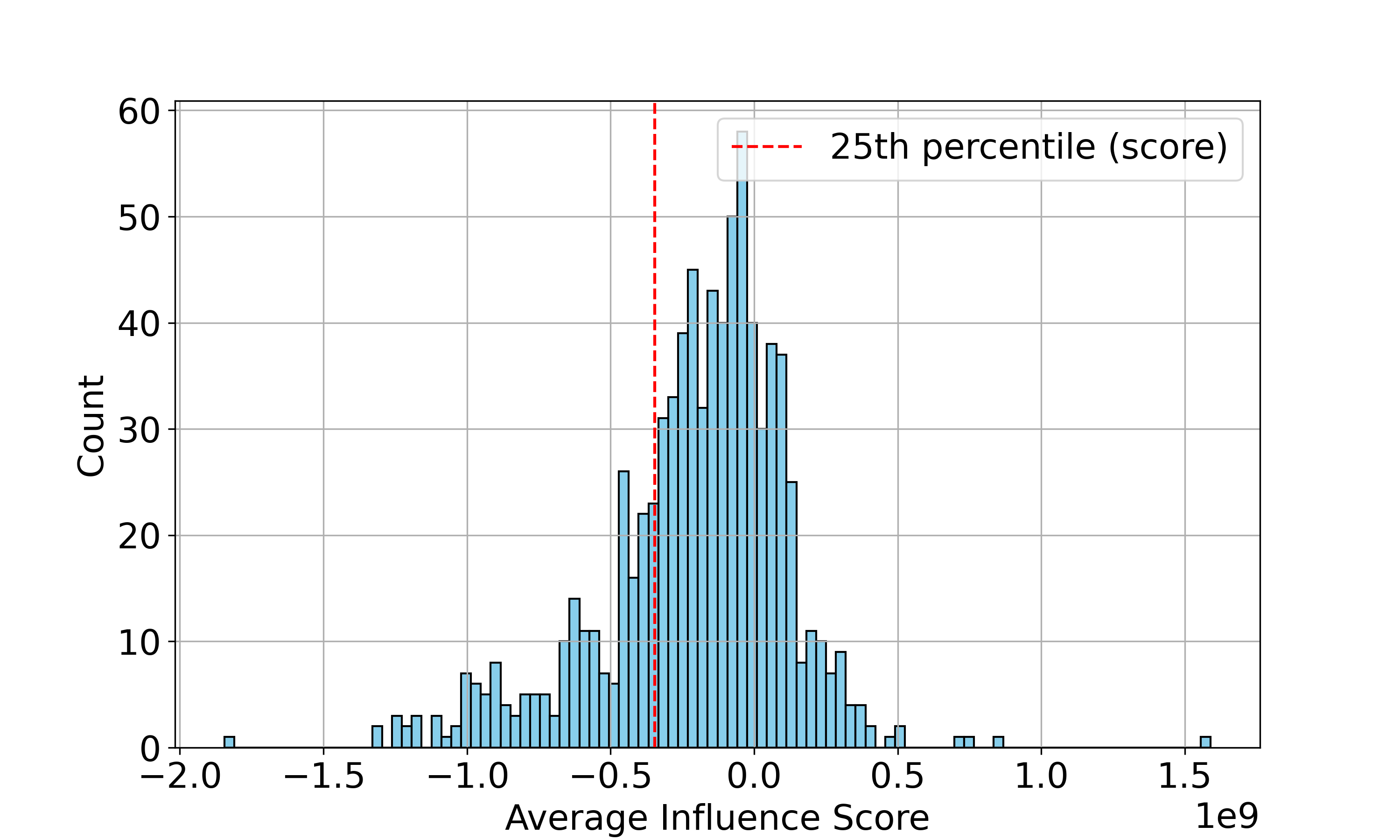}
    \label{fig:first}
\end{subfigure}
\vfill
\begin{subfigure}{\textwidth}
    \caption{\emph{Inorrect} influence score $s_I(d)$.}
    \includegraphics[width=\textwidth]{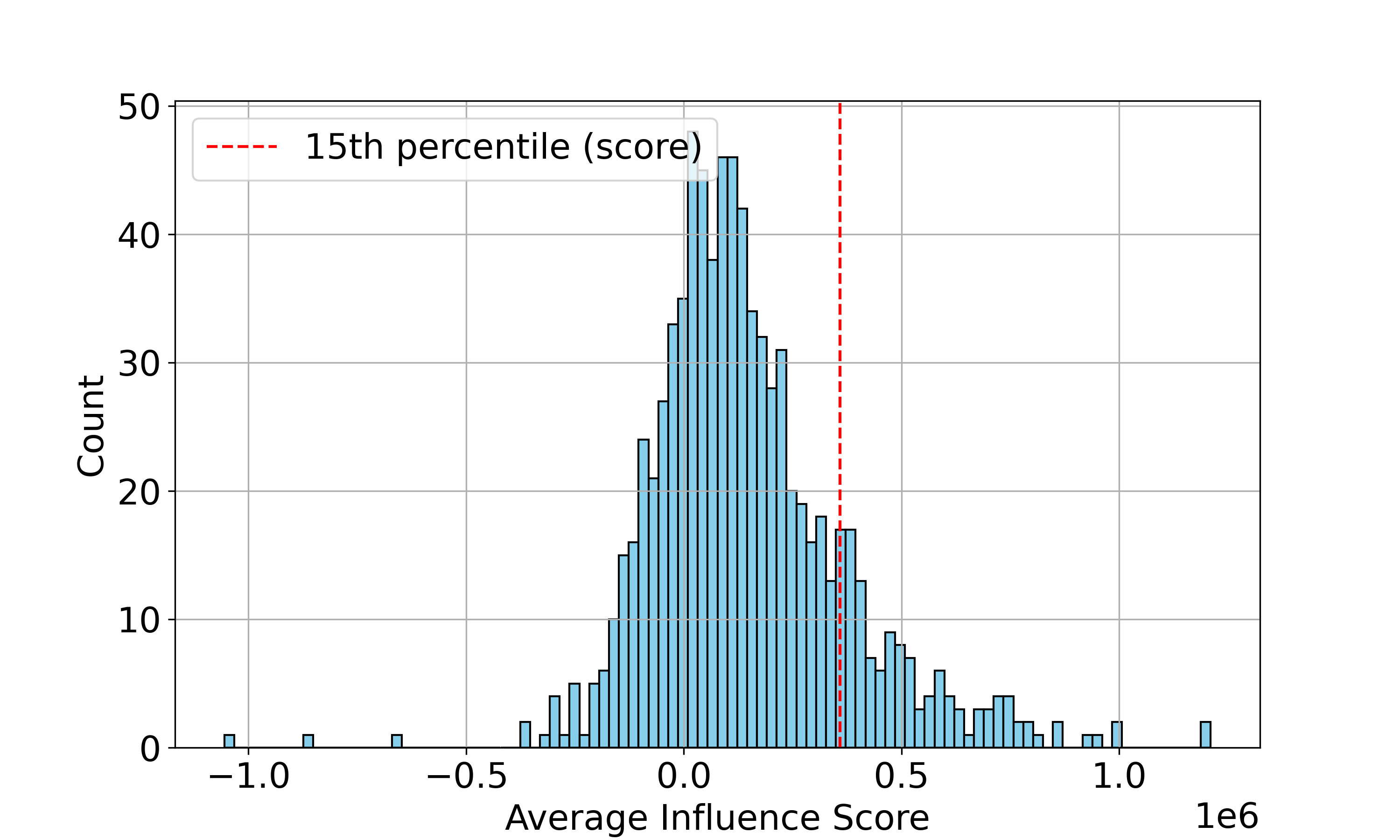}
    \label{fig:second}
\end{subfigure}
\end{minipage}
\begin{minipage}[b]{0.49\textwidth}
\begin{subfigure}{\textwidth}
    \caption{\emph{Correct} rank score $r_C(d)$.}
    \includegraphics[width=\textwidth]{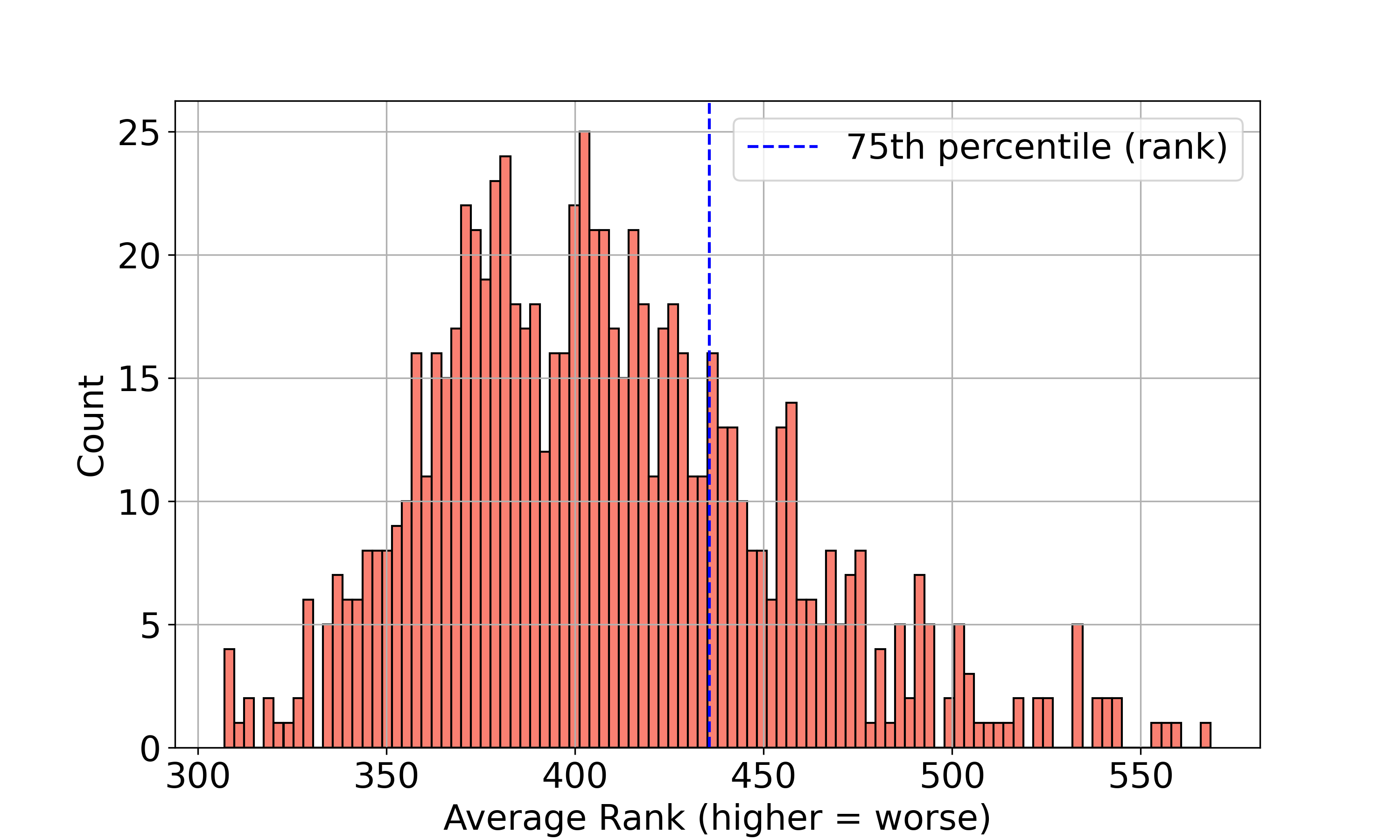}
    \label{fig:first}
\end{subfigure}
\vfill
\begin{subfigure}{\textwidth}
    \caption{\emph{Incorrect} rank score $r_I(d)$.}
    \includegraphics[width=\textwidth]{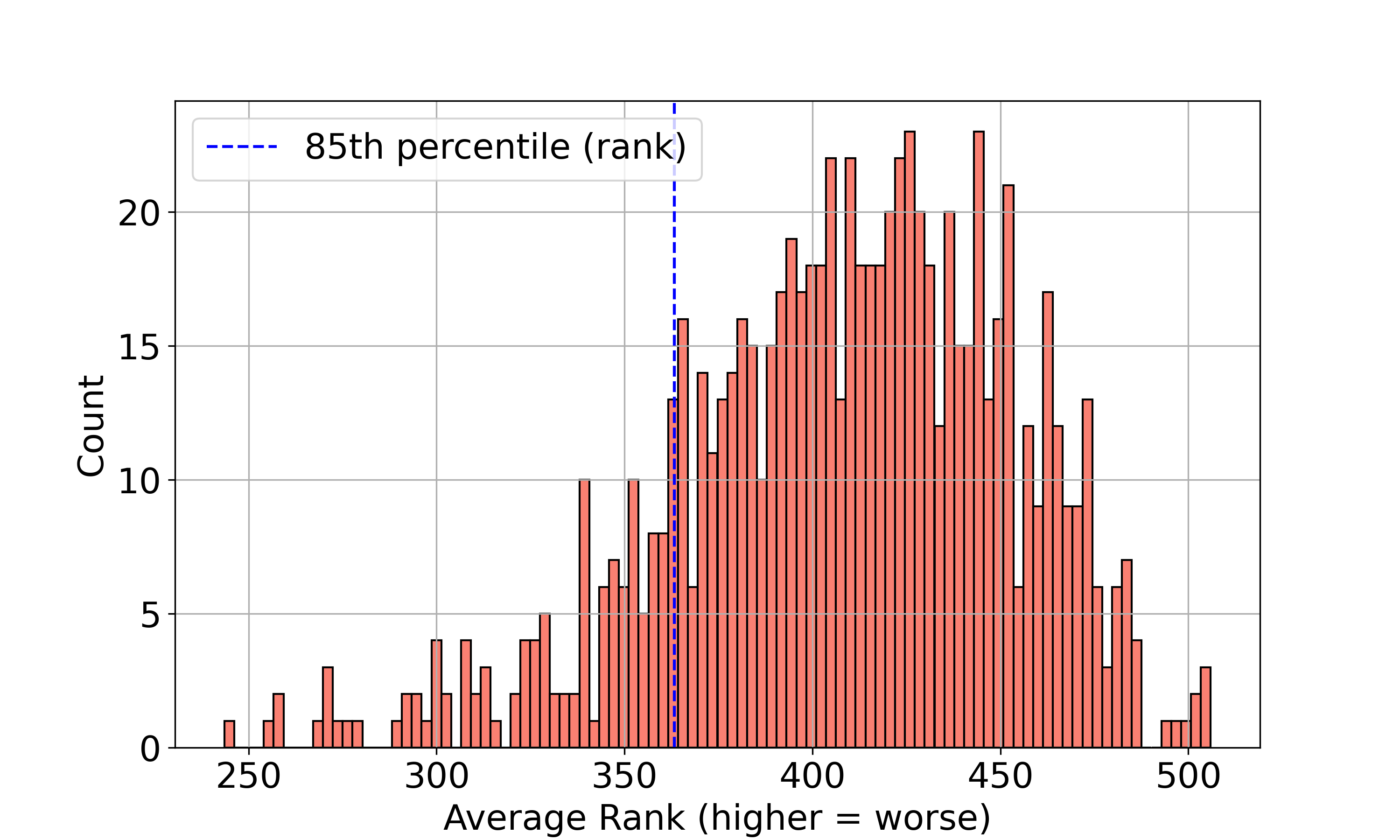}
    \label{fig:second}
\end{subfigure}
\end{minipage}

\caption{Histograms of the influence and rank scores measured across the training set $\mathcal{D}$.
}
\label{fig:histograms}
\end{figure}

In Figure~\ref{fig:histograms}, we report full details on the distribution of influence scores $s_C(d), s_I(d) $ and rank scores $r_C(d),r_I(d)$ over the over all data points $d \in \mathcal{D}$.
We notice how the distributions display long tails where we expect pruning to be more effective. 

\section{Results on Qwen2.5-Math-7B-Instruct}
We evaluate whether the dataset pruned using a specific model (LLaMA-3-8B-Instruct) would transfer its benefits when applied to a different model family. We report the full results on Qwen2.5-Math-7B-Instruct in Table~\ref{tab:main_res_qwen}.

\begin{table}[ht]
\centering
\caption{
\textbf{Results on \mbox{Qwen2.5-Math-7B-Instruct}.} Pass@1 with $N{=}8$ samples, zero-shot CoT. 
Same setup as Table~\ref{tab:main_res_llama}, but we fine-tune Qwen2.5-Math-7B-Instruct on pruned subsets selected using \mbox{LLaMA-3-8B-Instruct}, testing their transfer across model families. Best in \textbf{bold}. Amount of samples kept in [brackets].
}
\label{tab:main_res_qwen}
\vspace{.5em}
\footnotesize
\begin{tabular}{@{}lccccc@{}}
\multicolumn{6}{l}{\textsc{Qwen2.5-Math-7B-Instruct}}\\
\toprule
\textsc{Pruning} & \textsc{GSM8k} & \textsc{OlympiadBench} & \textsc{AMC23} & \textsc{AIME24} & \textsc{GPQA} \\
\midrule
Combined [90\%] (ours)   & 0.9567 & 0.4635 & 0.5781 & 0.1917 & 0.3220 \\
Incorrect [90\%] (ours)  & \textbf{0.9584} & 0.4643 & 0.5719 & 0.1667 & 0.3232 \\
Correct [80\%] (ours)    & 0.9566 & 0.4637 & 0.6000 & 0.1458 & 0.3245 \\
\cmidrule{1-6}
RDS+ [90\%]       & 0.9566 & 0.4669 & 0.6094 & 0.1583 & 0.3201 \\
Mid-PPL [90\%]    & 0.9571 & 0.4619 & \textbf{0.6125} & 0.1583 & 0.3169 \\
Random [90\%]     & 0.9575 & 0.4596 & 0.5781 & \textbf{0.2042} & 0.3295 \\
None              & 0.9576 & \textbf{0.4687} & 0.5469 & 0.1542 & \textbf{0.3340} \\
\cmidrule{1-6}
W/o SFT           & 0.9538 & 0.3876 & 0.5813 & 0.1125 & 0.2929 \\
\bottomrule
\end{tabular}
\end{table}

\section{More aggressive pruning (50\% of data)}
\subsection{Pruning strategy}
We experiment with a more aggressive IF-based pruning strategy, removing 50\% of the training data.

\begin{figure}[ht]
    \centering
    \includegraphics[width=0.5\linewidth]{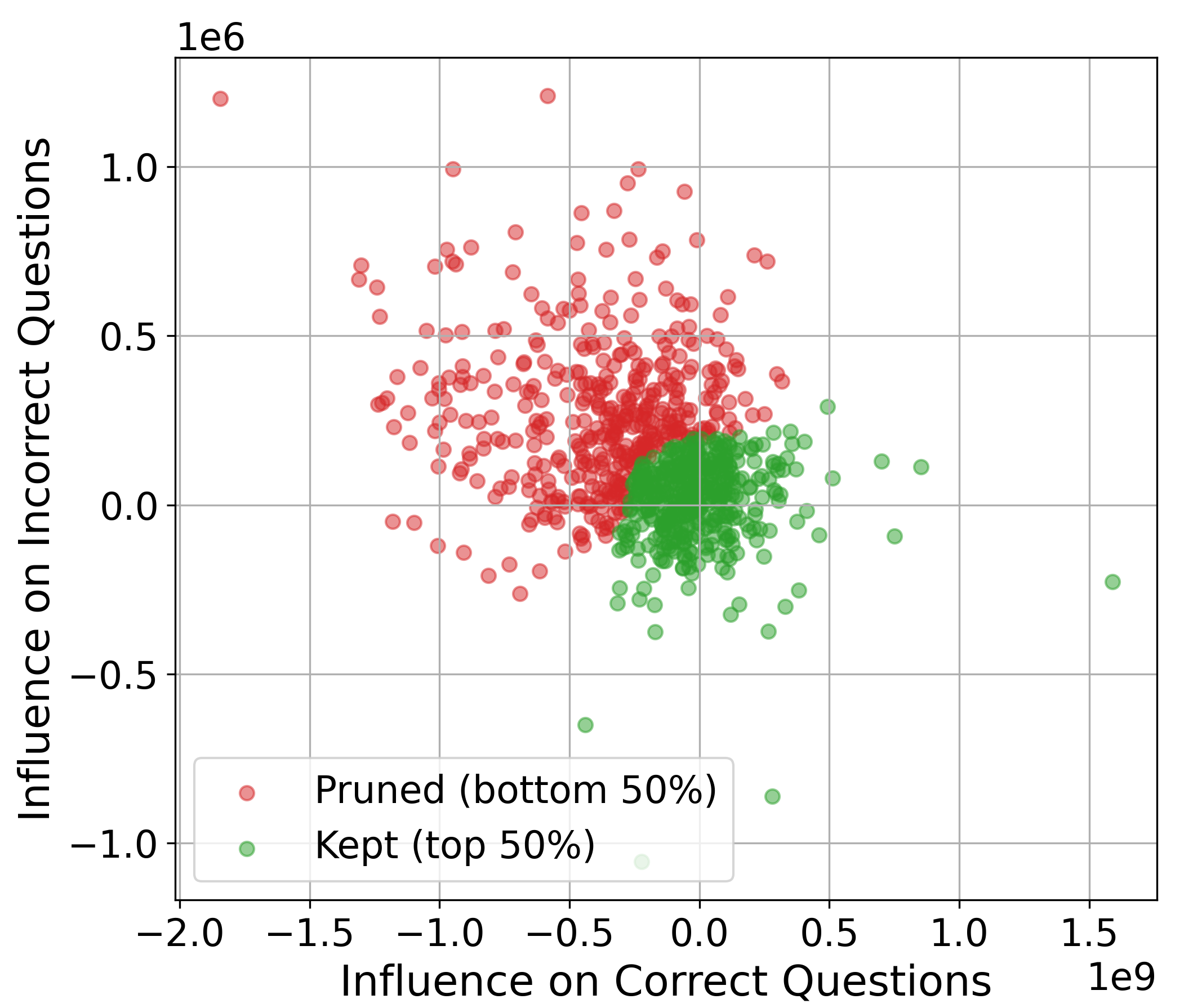}
    \caption{Influence scores $s_C(d), s_I(d)$ for training examples $d \in \mathcal{D}$ with aggressive 50\% data pruning strategy (\emph{Combined}).}
    \label{fig:scatter_50}
\end{figure}

Pruning is performed based on the following function:
$$A(s(v,d))=\frac{\text{sign}(s_C(d))s_C(d)^2}{\sigma_C} -\frac{\text{sign}(s_I(d))s_I(d)^2}{\sigma_I}, $$
taking the top 50\% of the resulting ranking, as illustrated also in Figure~\ref{fig:scatter_50}.

\subsection{Results}
The results for our \emph{Combined} strategy and all baselines 
confirm our main findings even under more aggressive pruning. 
As shown in Table~\ref{tab:res_50percent_llama}, IF-based pruning remains competitive with strong baselines: 
\emph{Combined (50\%)} is best on GSM8k, AMC23, and MATH500, while \emph{Random (50\%)} is slightly ahead on OlympiadBench.
On GPQA, performance improves compared to lighter pruning, confirming that reducing the amount of math-focused fine-tuning lessens the degradation of general scientific reasoning.

\begin{table}[ht]
\centering
\caption{\textbf{Results on \mbox{LLaMA-3-8B-Instruct}, pruning 50\% of the data.}
Pass@1 with $N{=}8$ samples, zero-shot CoT.
We fine-tune on math-pruned subsets selected by influence functions (\emph{Correct}, \emph{Incorrect}, \emph{Combined}) and compare against baselines (\emph{Random}, \emph{Mid-PPL}, \emph{RDS+}), as well as the unpruned dataset and base model. 
Evaluation covers math reasoning benchmarks (AIME24, AMC23, OlympiadBench, GSM8k) and GPQA for non-math transfer. Best in \textbf{bold}. Amount of samples not pruned in [brackets].}
\label{tab:res_50percent_llama}
\vspace{.5em}
\footnotesize
\begin{tabular}{@{}lcccc@{}}
\multicolumn{5}{l}{\textsc{Llama3-8B-Instruct}}\\
\toprule
\textsc{Pruning} & \textsc{GSM8k} & \textsc{OlympiadBench} & \textsc{AMC23} & \textsc{GPQA} \\
\midrule
Combined [50\%] (ours)  & \textbf{0.7981} & 0.0831 & \textbf{0.0969} & 0.1648 \\
\cmidrule{1-5}
RDS+ [50\%]        & 0.7962 & 0.0767 & 0.0813 & 0.1768 \\
Mid-PPL [50\%]     & 0.7797 & 0.0678 & 0.0938 & 0.1679 \\
Random [50\%]      & 0.7799 & \textbf{0.0843} & 0.0875 & \textbf{0.2134} \\
\cmidrule{1-5}
None        & 0.7722 & 0.0685 & 0.0813 & 0.1199 \\
W/o SFT     & 0.5958 & 0.0498 & 0.0781 & 0.2973 \\
\bottomrule
\end{tabular}
\end{table}

\section{Further plots}

\begin{figure}[ht]
\centering
\begin{subfigure}{0.48\linewidth}
  \includegraphics[width=\linewidth]{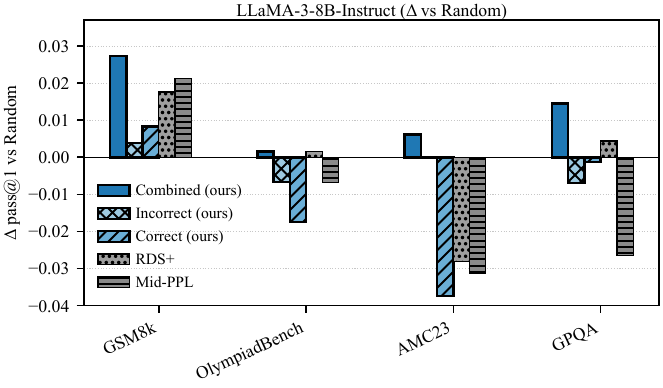}
  \caption{LLaMA-3-8B-Instruct}
\end{subfigure}
\hfill
\begin{subfigure}{0.48\linewidth}
  \includegraphics[width=\linewidth]{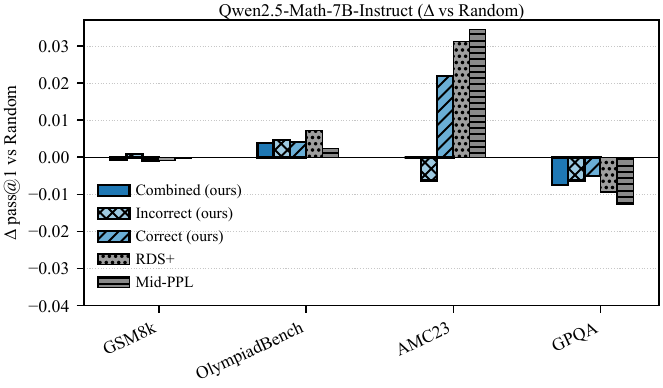}
  \caption{Qwen2.5-Math-7B-Instruct}
\end{subfigure}
\caption{
\textbf{$\Delta$pass@1 vs.\ Random pruning on benchmarks.}
\emph{(a)} For LLaMA-3-8B-Instruct, our IF-based Combined pruning achieves larger gains than baselines, particularly on GSM8k and OlympiadBench. On GPQA, \emph{Combined} mitigates this degradation more effectively than other pruning strategies.
\emph{(b)} For Qwen2.5-Math-7B-Instruct (~10\% pruning), improvements do not transfer reliably: all methods remain close to \emph{Random}, with AMC23 showing inconsistent deviations.
}
\label{fig:results_bar_delta_random}
\end{figure}

Figure~\ref{fig:results_bar_delta_random} visualizes how our pruning strategy outperforms other baselines for Llama3-8B-Instruct, while results are inconclusive on cross-family transfer with Qwen2.5-Math-7B-Instruct.

\end{document}